\documentclass[runningheads]{eccv_template/llncs}

\usepackage{eccv_template/eccv}

\usepackage{eccv_template/eccvabbrv}

\usepackage{graphicx}
\usepackage{booktabs}
\usepackage{xcolor}
\usepackage{colortbl}
\usepackage{siunitx}
\usepackage{makecell}
\usepackage{verbatim}
\usepackage{tabularx}
\usepackage{array}
\usepackage{arydshln}
\usepackage{multirow}
\usepackage{xspace}
\usepackage{marvosym} 
\usepackage{hyperref}

\usepackage{orcidlink}

\makeatletter
\newcommand{\printfnsymbol}[1]{%
  \textsuperscript{\@fnsymbol{#1}}%
}
\makeatother

\newcommand{\methodname}{\textsc{TensionTRAC}\xspace}
\newcommand{\dataname}{\textsc{SurgTension}\xspace}

\begin{document}

\title{Beyond Instrument Motion: Recognizing Tissue Tension Toward Surgical Skill Assessment}

\titlerunning{Beyond Instrument Motion: Tissue Tension Recognition}

\author{
Marko Haralovi\'c\thanks{Equal contribution, \textsuperscript{\Letter}Corresponding author.}\textsuperscript{,}\inst{1,2}    \orcidlink{0009-0004-1178-9964}
\and
Zhiqi Miao\printfnsymbol{1}\textsuperscript{,}\inst{3}\textsuperscript{,}\textsuperscript{\Letter} 
    \orcidlink{0009-0000-0043-7825}
\and
Alexander Machiel Bont\inst{4}
    \orcidlink{0009-0008-9922-7278}
\and
Jiapan Guo\inst{3}
    \orcidlink{0000-0003-3966-4405}
\and
Frans van Workum\inst{4,5}
    \orcidlink{0000-0003-0623-9066}
\and
Estefan\'ia Talavera\inst{2}
    \orcidlink{0000-0001-5918-8990}
}

\institute{
University of Zagreb, Faculty of Electrical Engineering and Computing, Zagreb, Croatia\\
\and
University of Twente, Enschede, The Netherlands\\
\and
University of Groningen, Bernoulli Institute, Groningen, The Netherlands\\
\and
Radboud University Medical Center, Nijmegen, The Netherlands\\
\and
Department of Surgery, Canisius-Wilhelmina Hospital, Nijmegen, The Netherlands\\
\email{marko.haralovic@fer.hr, z.miao@rug.nl}
}

\authorrunning{M.~Haralovi\'c, Z.~Miao et al.}

\maketitle


\begin{abstract}
Surgical performance assessment in minimally invasive surgery largely relies on manual expert review, making it time-consuming, subjective, and difficult to scale. While existing surgical video understanding methods address tasks such as instrument segmentation, surgical phase recognition, and action recognition, they do not explicitly capture fine-grained tissue handling, a key indicator of surgical quality. To address this gap, we introduce tissue tension recognition, a new clinically motivated video understanding task for laparoscopic and robot-assisted rectal cancer surgery. To support this task, we construct \dataname, the first clinically annotated and expert-reviewed tissue tension dataset, providing a benchmark for objective tissue tension recognition. We further propose \methodname, a lightweight trajectory-based framework that models tissue tension from sparse point trajectories. Using a compact trajectory encoder, \methodname achieves competitive performance against strong pretrained video backbones. Related material is available at
\href{https://github.com/MiaoMeow98/TensionTRAC}{Github - TensionTRAC}.

\end{abstract}

\begin{figure}[t]
    \centering
    \includegraphics[width=0.90\linewidth]{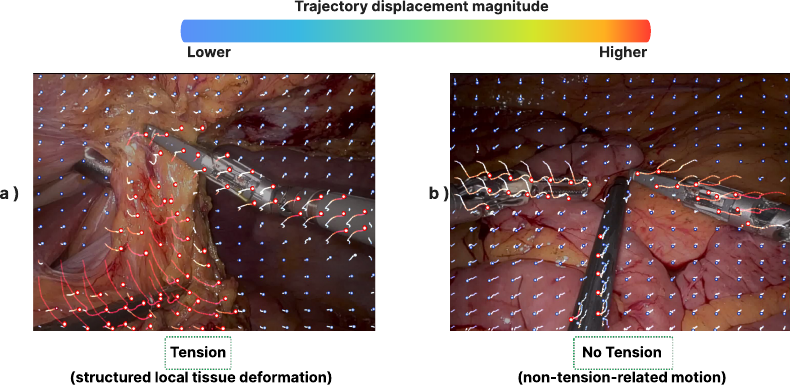}
\caption{Motivation for trajectory-based tissue tension recognition. If we uniformly sample points on a frame and employ a point tracker to track their movement over time (generating trajectories), we can observe that the tension-present clip (a) exhibits spatially structured local tissue deformation, whereas the no-tension clip (b) contains limited or tension-unrelated motion (e.g., instrument movement). This contrast motivates the use of sparse trajectory modeling for tissue tension recognition.}
    \label{fig:motivation}
\vspace{-5mm}
\end{figure}

\vspace{-6mm}
\section{Introduction}
\label{sec:intro}

Surgical performance is a key determinant of outcomes after complex procedures~\cite{curtis2020surgicalskill,ketel2024nationwide}. In minimally invasive surgery, including laparoscopic and robot-assisted rectal cancer surgery, the operative field is routinely recorded from the perspective of the surgeon. These videos provide an opportunity to assess technical performance after the procedure and to support surgical training and quality improvement~\cite{gruter2023video,ketel2024video}. However, in current clinical practice, surgical performance assessment still largely relies on manual review using competency assessment tools (CAT). Although such tools can provide reliable expert evaluation, the process is time-consuming and dependent on experienced surgical assessors~\cite{ketel2023cat,ketel2024video}.

\vspace{1.5mm}
Existing computer vision approaches for surgical video understanding have made substantial progress in tasks such as instrument segmentation~\cite{endovis2015,endovis2017robotic,endovis2018robotic,sarrarp50,sims2023,endovis2023}, tool presence detection~\cite{twinanda2016endonet,surgtooloc2022}, surgical phase and workflow recognition~\cite{endovis2017sensoror,heico,heichole,grasp,twinanda2016endonet,autolaparo,stauder2016lapchole,nephrec9,misaw}, and action recognition~\cite{cholectriplet2021,cholectriplet2022,cholect40,cholect50,sarrarp50,psiava,esad,jigsaws,lsrao,lin2024instrument}. These tasks capture important aspects of the surgical scene, but \textbf{these tasks alone do not fully reflect how surgeons evaluate operative quality}. Surgical expertise is often expressed through fine-grained technical behaviors, such as tissue handling, exposure, traction and counter-traction, and safe dissection. These behaviors are not always well described by phase labels, tool categories, or global video-level action representations. Therefore, as current tasks do not correspond well to how
surgeons view surgical quality, there is a need to move beyond 
surgical video recognition tasks and model clinically meaningful cues that are directly related to surgical quality.

\vspace{1mm}
In this work, we focus on one cue: \textbf{tissue tension}. During rectal cancer surgery, dissection is performed in a narrow anatomical space, where appropriate traction and counter-traction are essential for exposing the correct surgical dissection plane. Insufficient tissue tension can reduce visibility and surgical efficiency, while excessive tension may increase the risk of tissue damage. In contrast, adequate or clinically acceptable tension reflects a desirable operating condition in which the tissue is sufficiently exposed without being over-stretched. Despite its clinical relevance, to the best of our knowledge, tissue tension has not been systematically studied as a computer vision task and there is currently no established benchmark or automatic method for recognizing tissue tension from minimally invasive surgical videos. 

A central challenge is that \textbf{tissue tension is not simply a static appearance category, but a dynamic state that emerges from local instrument--tissue interactions and evolves over time}. Such temporally varying patterns may not be fully captured from the global surgical scene alone. Standard dense video representations, including CNN- and Transformer-based encoders~\cite{ji20133d,feichtenhofer2019slowfast,bertasius2021timesformer,tong2022videomae,liu2022videoswin,wu2026surgmotion}, typically aggregate appearance and motion information across the entire frame. Although effective for surgical phase and action recognition, these representations may be influenced by scene context, instrument appearance, and background structures, and may therefore underemphasize the sparse local motion and tissue deformation cues that indicate whether tissue is being pulled, stabilized, or released~\cite{kumar2024tats,kumar2025trokens,meier2026traction}.

We formulate tissue tension recognition as a sparse trajectory-based problem that not only focuses on clinically relevant local motion cues but also avoids the heavy computation associated with dense video modeling. Instead of relying solely on coarse video-level or dense frame-level features, we track sparse visual points over time and use their trajectories to characterize local tissue motion, deformation, and instrument-tissue interaction. This point tracking paradigm is particularly suitable for tissue tension recognition, as shown in Fig~\ref{fig:motivation}, because tension is expressed through how local regions move, stretch, remain stable, or relax across time. By shifting from global video representation to sparse point trajectory modeling, our approach explicitly focuses on the temporal dynamics that are clinically relevant for tension estimation. To support this problem formulation, we introduce an clinically annotated protocol for tissue tension in real world da Vinci robot-assisted rectal cancer surgery videos. The annotations include different tension states, allowing to study both the distribution of tissue tension events and the feasibility of automatic tension recognition. This work provides an initial step toward trajectory-based tissue tension recognition, with the long-term goal of supporting clinically meaningful assessment of surgical performance from minimally invasive surgical videos.

Our contributions are threefold. 
(1) We introduce \textbf{tissue tension recognition} as a new clinically motivated task for surgical video understanding, targeting \textbf{fine-grained tissue handling quality assessment} beyond conventional phase-, action-, and instrument-centric analysis (Section \ref{sec:tension_task_dataset}). 
(2) We present a clinically annotated and expert-reviewed \textbf{tissue tension dataset \dataname} constructed from real-world rectal cancer surgery videos. The annotations and releasable data will be made publicly available to facilitate future research (Section \ref{sec:tension_task_dataset}). 
(3) We propose \methodname, a \textbf{lightweight sparse point tracking-based framework} for tissue tension recognition that shifts the representation paradigm from \textbf{dense video-level appearance modeling} to \textbf{trajectory-based modeling of local tissue and instrument--tissue dynamics}, while achieving competitive performance against strong video-based baselines (Section~\ref{sec:methodology}).

\section{Related Work}
\label{sec:related_work}

\noindent\textbf{Surgical video understanding and datasets.}
Surgical video analysis has been explored across laparoscopic surgery~\cite{twinanda2016endonet,stauder2016lapchole,heico,heichole}, surgical training~\cite{jigsaws,misaw}, robot-assisted surgery on human subjects~\cite{nephrec9,lsrao,esad,sarrarp50,autolaparo,rarr,psiava,grasp}, and animal or phantom settings~\cite{endovis2015,endovis2017robotic,endovis2018robotic,surgtooloc2022,endovis2017sensoror,endovis2023}. Existing datasets have enabled progress on multi-level surgical scene understanding, including phase recognition~\cite{twinanda2016endonet,stauder2016lapchole,cholect40,cholect50,heico,misaw,autolaparo,rarr,heichole,psiava,grasp}, step recognition~\cite{nephrec9,misaw,psiava,grasp}, instrument recognition, detection and segmentation~\cite{endovis2015,endovis2017robotic,endovis2018robotic,heico,sarrarp50,autolaparo,grasp}, action recognition~\cite{jigsaws,endovis2018robotic,lsrao,cholect40,cholect50,misaw,esad,sarrarp50,heichole,psiava,grasp}, and action localization~\cite{endovis2018robotic,esad,psiava,grasp}. These tasks capture important workflow, tool, and action information, but their annotations primarily describe what is happening in the surgical scene, rather than how tissue is being handled. In particular, existing benchmarks rarely provide event-level labels indicating whether tissue traction is insufficient, clinically acceptable or excessive during dissection. In contrast, our work focuses on tissue tension recognition in da Vinci robot-assisted laparoscopic rectal cancer surgery and introduces a clinically annotated dataset with event-level tissue tension labels.

\noindent\textbf{Surgical quality assessment and tissue handling.}
Manual surgical performance assessment is commonly performed using structured competency assessment tools. The competency assessment tools (CATs) provide procedure-specific frameworks for expert evaluation of technical surgical performance~\cite{curtis2020surgicalskill,ketel2023cat,ketel2024video}. Such tools are designed to support structured feedback, surgical training, and performance evaluation, but manual video review remains time-consuming and difficult to scale. Importantly, CAT-based assessment often reflects fine-grained technical behaviors, such as tissue handling, exposure, traction, counter-traction, and safe dissection, which are not directly measured by conventional phase, tool, or action labels. Related work has explored vision-based surgical force estimation~\cite{jung2020vision,wang2026exploring}. However, these methods generally estimate continuous force-related signals from visual observations, whereas we formulate tissue tension as an event-level surgical video recognition task with clinically defined states. To the best of our knowledge, no established event-level benchmark exists for tissue tension recognition in minimally invasive rectal cancer surgery. We therefore introduce this task together with a clinically annotated dataset of da Vinci robot-assisted rectal cancer surgery videos, focusing on total mesorectal excision.

\noindent\textbf{Video representations and trajectory modeling.}
Existing surgical video recognition methods can be broadly grouped into frame-based 2D CNN approaches~\cite{twinanda2016endonet,cholect40,surgtooloc2022}, CNNs with temporal sequence modeling~\cite{lsrao,misaw}, spatiotemporal CNN video encoders~\cite{feichtenhofer2019slowfast,lsrao}, and transformer-based video models~\cite{bertasius2021timesformer,cholect50,psiava,grasp}. These models are effective for many surgical recognition tasks, but they typically rely on dense RGB frame-level or clip-level feature aggregation and do not explicitly capture local tissue deformation and instrument-tissue interactions. Furthermore, their computational demands can limit scalability in real-world surgical applications, motivating lightweight representations that directly model relevant motion patterns. Recent point tracking and trajectory-based video representations have demonstrated the potential of sparse motion modeling for dynamic scene understanding and action recognition~\cite{karaev2024cotracker, kumar2024tats,kumar2025trokens}. However, their potential for capturing fine-grained tissue dynamics in surgical videos remains unexplored. To address this gap, we propose a sparse point tracking-based framework for tissue tension recognition that models temporally evolving tissue motion and deformation.
\section{Tissue Tension Recognition: \dataname Dataset and Tasks}
\label{sec:tension_task_dataset}

\subsection{Clinical Definition of Tissue Tension}
\label{sec:tension_definition}

In this work, \textit{tissue tension} refers to the visually apparent mechanical state produced when a surgeon applies traction or counter-traction before or during dissection. In robot-assisted rectal surgery, appropriate tension exposes anatomical landmarks and opens the intended dissection plane, thereby facilitating controlled dissection or cutting. In contrast, insufficient, excessive, or incorrectly directed traction may impair exposure, obscure or collapse the dissection plane, and increase the risk of tissue injury, bleeding, avulsion, or loss of the correct anatomical plane. This definition is informed by CAT-based surgical performance assessment, which considers domains such as exposure, task execution, errors, and end-product quality. Within these domains, expert performance is associated with clear exposure, appropriate traction and counter-traction, atraumatic tissue handling, dissection under suitable tension, and preservation of the correct anatomical plane. We therefore formulate tissue tension recognition as a surgical video perception task in which short clips are categorized according to the visually apparent tension level and resulting tissue deformation, with emphasis on their relevance to exposure and dissection.

\subsection{\dataname Dataset Collection and Preprocessing}
\label{sec:dataset}

\dataname was constructed from seven robot-assisted rectal cancer resection videos recorded during routine clinical care at a teaching hospital using the da Vinci surgical platform. Before transfer for technical analysis, the recordings were de-identified locally and assigned unique study identifiers. The technical team received no identifiable patient information and could not link the recordings to patient records. In accordance with the institution's opt-out policy, only recordings from patients who had not opted out were included. The data are stored in encrypted research environments with access restricted to authorized personnel, in accordance with institutional data-governance procedures and applicable data-protection regulations. The recordings originated from routine clinical care and involved no study-specific intervention. The dataset comprises 11 hours, 7 minutes, and 44 seconds of video processed at 30 fps. Preprocessing included removing black borders and lower system overlays containing robot-system and instrument-status information.

\subsection{Annotation Protocol and Label Space}
\label{sec:annotation}

Annotations are performed by a sixth-year medical student with clinical experience in surgery under the direct supervision of a gastrointestinal surgeon with 15 years of clinical experience. Ambiguous cases are jointly reviewed with the supervising surgeon and resolved through consensus discussion. We therefore describe \dataname as a clinically annotated and expert-reviewed dataset. As the annotations are not independently performed by multiple experts, inter-rater reliability could not be quantified; independent multi-expert validation remains an important direction for future work. Using an in-house annotation tool, the annotators mark the start and end timestamps of each tension event and assign the corresponding approximately 1-second clip to one of seven categories: \textit{non-tension}, \textit{low tension}, \textit{moderate tension}, \textit{adequate tension}, \textit{excessive tension}, \textit{reverse tension}, or \textit{unclear tension}. Their operational definitions and distributions are provided in Table~\ref{tab:tension_cat_desc}. \textit{Reverse tension} indicates complete tension reduction or counterproductive traction, whereas \textit{unclear tension} denotes clips that cannot be assessed reliably; both are excluded. In total, 3,867 clips are annotated. After excluding 208 \textit{reverse-tension} and 66 \textit{unclear} clips, the final benchmark contains \textbf{3,593 clips}: 874 \textit{non-tension} clips (level 0) and 2,719 tension-present clips spanning \textit{low}, \textit{moderate}, \textit{adequate}, and \textit{excessive tension} (levels 1--4).

\begin{table}[!t]
\centering
\scriptsize
\caption{\dataname tissue tension dataset categories and category distribution.}
\label{tab:tension_cat_desc}
\setlength{\tabcolsep}{4pt}
\renewcommand{\arraystretch}{1.05}
\begin{tabularx}{\linewidth}{@{} l X r r @{}}
\toprule
\textbf{Category} & \textbf{Clinical description} & \textbf{\#} & \textbf{\%} \\
\midrule
0 Non-tension ($\checkmark$) & No visible traction-induced tissue tension or deformation. & 874 & 22.6 \\
1 Low ($\checkmark$) & Minimal traction or tissue deformation, with no apparent risk of tissue injury. & 545 & 14.1 \\
2 Moderate ($\checkmark$) & Visible traction and tissue deformation sufficient to facilitate exposure or dissection. & 1349 & 34.9 \\
3 Adequate ($\checkmark$) & Pronounced but controlled traction and tissue deformation within the clinically acceptable range. & 749 & 19.4 \\
4 Excessive ($\checkmark$) & Marked or uncontrolled tissue stretching beyond the clinically acceptable range, with increased risk of tissue injury or loss of the dissection plane. & 76 & 2.0 \\
\midrule
Reverse ($\times$) & Reduction, release, or reversal of previously applied tissue tension. & 208 & 5.4 \\
Unclear ($\times$) & Insufficient visibility, occlusion, poor image quality, or uncertainty preventing reliable assessment. & 66 & 1.7 \\
\midrule
\textbf{Total annotated} & & \textbf{3867} & \textbf{100} \\
\textbf{Final valid clips} & & \textbf{3593} & \textbf{93} \\
\bottomrule
\end{tabularx}
\vspace{-4mm}
\end{table}

\subsection{Tissue Tension Recognition as a Video Understanding Task}
\label{sec:task_formulation}

Given a preprocessed surgical clip $C$, the goal of tissue tension recognition is to predict a tension label $\hat{y}$ derived from surgeon annotations. Based on the label space defined in Sec.~\ref{sec:annotation}, we formulate two complementary tension recognition tasks that share the same annotated clips but differ in the clinical granularity of the prediction target.

1) \textbf{Binary tension classification}:
The binary task distinguishes \textit{non-tension} from \textit{tension-present} clips. Level 0 is mapped to the negative class, while all levels 1--4 are mapped to the positive class. This formulation evaluates whether a model can detect the presence of traction-induced tissue deformation and is defined over all 3593 valid clips.

2) \textbf{Cascade tension classification}:
The cascade task follows the more realistic clinical decision process. Stage 1 first determines whether tension is present. Stage 2 then grades the tension level for tension-present clips. The final end-to-end output spans levels 0--4, with level 0 corresponding to non-tension and levels 1--4 corresponding to increasing tension levels. This formulation evaluates both tension detection and tension grading in a unified setting over all valid clips.

The benchmark also supports two complementary assessment granularities: Clip-level stratified evaluation and cross-video grouped evaluation. In the clip-level setting, each annotated clip is treated as an independent sample, directly measuring recognition performance on localized surgical moments. In the video-level setting, clips are grouped according to their source surgical video, allowing evaluation under a more clinically realistic scenario where models must generalize across procedures rather than rely on visually similar clips from the same recording. The details of evaluation protocols is described in the Section.~\ref{sec:experiments}.

\section{\methodname: Trajectory-Based Tension Recognition}
\label{sec:methodology}
To model tissue tension dynamics in surgical videos, we first motivate the use of trajectory-based representations for capturing fine-grained motion patterns and long-term temporal evolution in complex surgical scenarios. Unlike conventional video representations that rely primarily on dense appearance modeling, trajectories provide a compact description of local tissue movements, deformation patterns, and instrument--tissue interactions over time. Based on this motivation, we propose \textbf{\methodname}, a trajectory-based framework for tissue tension recognition. It represents each surgical clip using sparse point trajectories together with their associated semantic features. We leverage DINOv2 patch features~\cite{oquab2023dinov2} to encode the rich semantic appearance of tracked tissue regions, instruments, and surrounding surgical context, and combine them with trajectory-derived motion descriptors to predict the tissue tension state.

\subsection{Why Trajectories for Tension Recognition?}
\label{sec:why_trajectories}

Tissue tension is inherently a motion- and interaction-driven phenomenon. Unlike many surgical scene understanding tasks that can be inferred primarily from static appearance, tension arises when an instrument applies traction or counter-traction to surrounding tissue, causing visible deformation, displacement, stretching, and coordinated motion between tissue regions. These cues are often subtle, local, and temporally expressed: the same tissue region may appear visually similar before and after traction, while its motion pattern can change substantially once force is applied. This makes point trajectories a natural representation for capturing the visual dynamics of tissue tension.

Thanks to recent advances in point-tracking foundation models~\cite{karaev2024cotracker,karaev2025cotracker3}, several trajectory-based action recognition methods have emerged and demonstrated that explicit motion traces can provide strong representations for human-object and object-object interactions~\cite{kumar2024tats,kumar2025trokens,meier2026traction}. Inspired by this observation, we investigate whether tissue tension also induces measurable trajectory-level signatures in surgical videos. Beyond that, sparse trajectories provide a compact representation of surgical motion with low computational cost. By modeling only a limited set of tracked points rather than dense spatio-temporal tokens, sparse trajectory representations can largely reduce the number of tokens and input frames required relative to dense video encoders while retaining motion cues relevant to tissue manipulation. This efficiency advantage is particularly attractive for surgical video analysis, where long procedures and high-resolution recordings can make dense spatio-temporal modeling computationally expensive.

We use a frozen foundation point tracker~\cite{karaev2025cotracker3} with uniformly sampled points on tension and non-tension clips, and compute trajectory-derived statistics including visibility, velocity, acceleration, displacement change, stretch, strain energy motion, motion coherence, and spatial displacement. To examine whether trajectory descriptors are informative for tension recognition, we perform a two-sided statistical test for each trajectory metric between tension and non-tension clips. As shown in Table~\ref{tab:traj_motivation}, $10/14$ trajectory metrics differ significantly at the nominal level of $\alpha=0.05$, indicating that tension and non-tension clips exhibit distinct trajectory-level patterns, as reflected by these statistics. Importantly, the significant differences are consistent with the expected mechanics of tissue tension: tension clips show stronger motion, larger inter-point distance variation, greater stretch, and higher strain energy deformation. These results suggest that tissue tension is associated with measurable changes in sparse trajectory dynamics. The significant differences in motion intensity, inter-point distance variation, stretch, and strain energy motion indicate that trajectories capture tension-related cues that are difficult to characterize from static appearance alone. This motivates our trajectory-based formulation for tissue tension recognition.

\begin{table}[h]
\centering
\footnotesize
\scriptsize
\setlength{\tabcolsep}{8.0pt}
\renewcommand{\arraystretch}{1.02}
\begin{tabular}{lrrrr}
\toprule
\textbf{Metric} & \textbf{Non-Tension} & \textbf{Tension} & \textbf{$\Delta$} & \textbf{$p$} \\
\midrule
Visible pts.      & 87.11  & 77.33  & -9.78   & $\mathbf{1.06{\times}10^{-8}}$ \\
Velocity mean     & 1.78   & 2.29   & +0.51   & $\mathbf{1.11{\times}10^{-5}}$ \\
Velocity max      & 27.84  & 30.90  & +3.06   & $\mathbf{7.60{\times}10^{-3}}$ \\
Accel. mean       & 0.66   & 0.69   & +0.03   & $\mathbf{2.40{\times}10^{-2}}$ \\
Accel. max        & 15.31  & 11.76  & -3.55   & $0.633$ \\
Dist. change mean & 0.23   & -0.30  & -0.53   & $0.209$ \\
Dist. std. time   & 13.13  & 18.43  & +5.29   & $\mathbf{8.69{\times}10^{-7}}$ \\
Max stretch       & 2.87   & 3.49   & +0.62   & $\mathbf{2.34{\times}10^{-4}}$ \\
Mean max stretch  & 1.04   & 1.06   & +0.02   & $\mathbf{1.34{\times}10^{-4}}$ \\
Strain E. mean    & 0.0225 & 0.0372 & +0.0147 & $\mathbf{1.78{\times}10^{-5}}$ \\
Strain E. max     & 0.0644 & 0.0810 & +0.0166 & $\mathbf{1.13{\times}10^{-3}}$ \\
Motion coh. mean  & 0.672  & 0.711  & +0.039  & $0.0522$ \\
Spatial disp. mean& 470.03 & 462.02 & -8.01   & $\mathbf{8.36{\times}10^{-5}}$ \\
Spatial disp. max & 487.72 & 487.19 & -0.53   & $0.249$ \\
\bottomrule
\vspace{2mm}
\end{tabular}
\caption{
Trajectory-derived motion and deformation statistics for Non-Tension and Tension clips. 
For each metric, Non-Tension and Tension report the average value over clips, and $\Delta=\text{Tension}-\text{Non-Tension}$ indicates the magnitude of the difference. 
The $p$ value is obtained from a two-sided statistical test comparing the two groups. 
Significant differences at the nominal level of $\alpha=0.05$ are bolded. 
The results show that tension clips exhibit stronger motion and deformation patterns, supporting the use of sparse trajectories for tissue tension recognition.
}
\label{tab:traj_motivation}
\end{table}

\subsection{\methodname: Trajectory-Based Tissue Tension Recognition}
\begin{figure}[h!]
    \centering    \includegraphics[width=\linewidth]{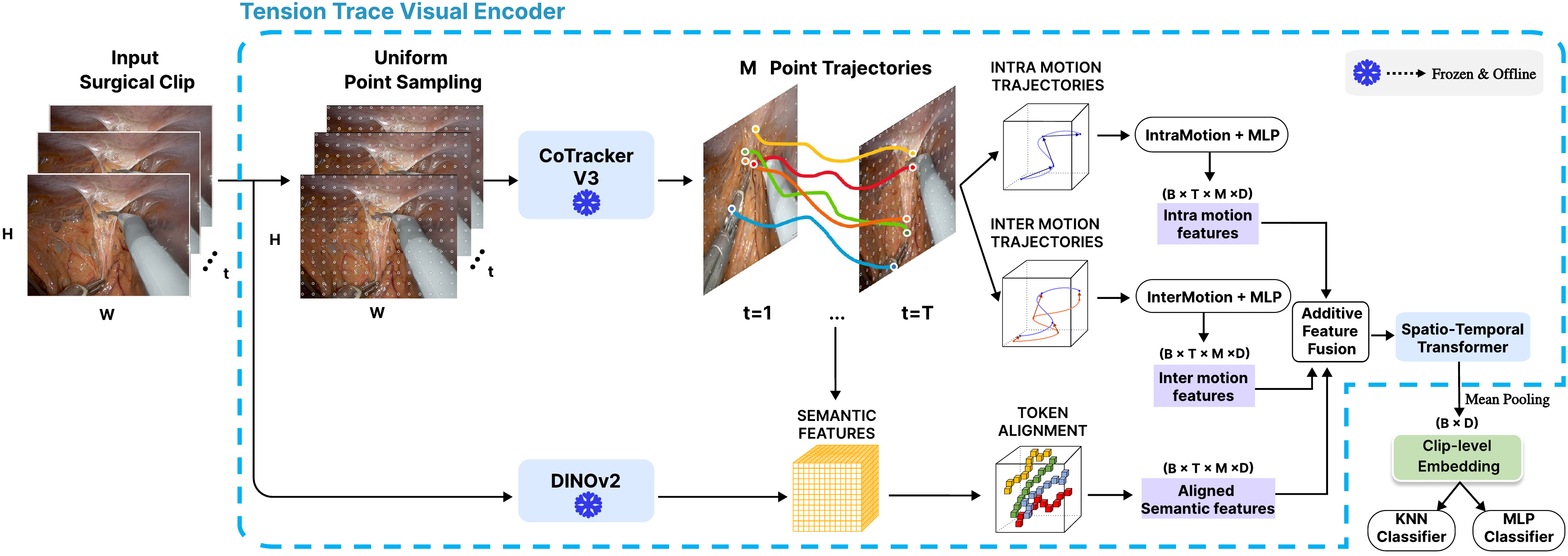}
    \caption{Overview of \methodname. Frozen DINOv2 and CoTrackerV3 models are used offline for semantic-feature extraction and trajectory generation, respectively. The intra-trajectory and inter-trajectory motion branches, feature fusion, and spatio-temporal transformer form the pretrained trajectory encoder, which is frozen to extract clip-level representations for downstream $k$-NN and MLP classification on SurgTension.}
    \label{fig:trajectory_action_classifier}
\vspace{-2mm}
\end{figure}

For each clip $C$ with $T$ frames, we extract frame-wise semantic features using a frozen DINOv2~\cite{oquab2023dinov2} visual encoder and initialize a set of sparsely sampled points on a uniform $n \times n$ grid in the first frame. These points are tracked across the clip with a frozen foundation point tracker, producing a set of trajectories that provide sparse yet informative motion cues about how tissues and instruments move over time. For each trajectory, we encode intra-trajectory dynamics with histogram-based descriptors computed from successive 2D displacements. We further model inter-trajectory interactions by encoding pairwise relative positions among trajectories at each time step, producing intra- and inter-trajectory motion features. In parallel, we sample the backbone's semantic features at the tracked point locations in each frame, obtaining point-aligned appearance features. All of these features are fused and processed by a one-block spatio-temporal transformer. Finally, the transformer outputs are aggregated by visibility-weighted mean pooling to obtain a clip-level representation for tissue tension classification. An overview of the framework is shown in Figure~\ref{fig:trajectory_action_classifier}.

\paragraph{\textbf{(1) Visual Encoder: Trajectory-based Representation}}

\paragraph{\textbf{Uniform Point Sampling.}}
A set of $M$ sparse points is initialized on a uniform $n\times n$ grid in the first frame of each clip. 
These points are then tracked across the clip using the frozen CoTracker~\cite{karaev2025cotracker3} model, producing a set of point trajectories 
$\mathcal{P}=\{\mathcal{P}^{m}\}_{m=1}^{M}$.
Each trajectory is defined as $\mathcal{P}^{m}=[(x_t^m,y_t^m)]_{t=1}^{T}\in\mathbb{R}^{T\times2},$
where $x$ and $y$ denote the image-plane coordinates. Therefore, each trajectory records the image-plane motion of one sampled point over time.

\paragraph{\textbf{Trajectory-aligned Semantic Features.}}
For each frame in a clip, we extract DINOv2~\cite{oquab2023dinov2} patch features and sample semantic descriptors at each tracked point location across frames, producing trajectory-aligned appearance features $\mathbf{F}^{\mathrm{sem}} \in \mathbb{R}^{B \times T \times M \times D}$.

\paragraph{\textbf{Intra-trajectory Motion Feature.}}
For each tracked point in each trajectory, we compute 2D histogram-of-displacements (HoD)~\cite{gowayyed2013hod} feature, preserving the temporal order of the displacement statistics. Let the HoD descriptor be
\begin{equation}
\mathbf{H}_{\mathrm{HoD}}^{\mathrm{intra}} = f_\text{HoD}(\mathcal{P}^m) \in \mathbb{R}^{T \times K},
\label{eq:hod}
\end{equation}
where $K$ is the number of bins. The HoD descriptor is then projected by an MLP into the model feature space to produce intra-trajectory motion features.

\begin{equation}
\mathbf{F}_{\text{intra}}^{\text{motion}}
= \operatorname{MLP}\left(
\mathbf{H}_{\mathrm{HoD}}^{\mathrm{intra}}
\right)
\in \mathbb{R}^{B \times T \times M \times D},
\label{eq:intra_motion_traj}
\end{equation}

\paragraph{\textbf{Inter-trajectory Motion Feature.}}
To model cross-point interactions, we compute pairwise relative coordinates between each trajectory $m$ and all other trajectories at the same time step:

\begin{equation}
\mathbf{r}_t^m = \big[\mathbf{p}_t^m - \mathbf{p}_t^1,\ldots,\mathbf{p}_t^m - \mathbf{p}_t^M\big] \in \mathbb{R}^{2M},
\label{eq:cross_motion}
\end{equation}
with invalid positions masked by trajectory visibility. The resulting descriptor is projected by an MLP to obtain inter-trajectory motion features
\begin{equation}
\mathbf{F}_{\mathrm{inter}}^{\mathrm{motion}}=\operatorname{MLP}(r)\in \mathbb{R}^{B \times T \times M \times D}.
\label{eq:inter_motion_traj}
\end{equation}

\paragraph{\textbf{Spatio-Temporal Transformer.}}
The $\mathbf{F}^{\mathrm{sem}}$, $\mathbf{F}_{\text{intra}}^{\text{motion}}$, and $\mathbf{F}_{\text{inter}}^{\text{motion}}$ are projected to a shared feature dimension and fused by addition. The fused representation is processed by a one-block spatio-temporal transformer. Specifically, temporal attention is first applied independently along the $T$ frames of each trajectory to model how the motion and appearance of a single track evolve over time. Spatial attention is then applied at each frame across the $M$ trajectories to model instantaneous interactions among co-occurring tracks. The resulting clip-level representation is used for subsequent tissue tension recognition.

\paragraph{\textbf{(2) Tension Classification}}
We evaluate all representations with a frozen-feature protocol. For each clip $C$, the visual encoder produces a fixed embedding $\mathbf{z}=f(C)$. For \methodname, $f(\cdot)$ denotes the pretrained tension trace visual encoder followed by visibility-weighted mean pooling over the spatio-temporal point tokens; for video baselines, $f(\cdot)$ is the pretrained video backbone encoder. 

We adopt two frozen-feature classifiers. First, a non-parametric $k$-nearest-neighbor classifier is trained on the training folds. At test time, each query embedding $\mathbf{z}_q$ is compared with training embeddings using cosine similarity after per-fold standardization and $L_2$ normalization. The label is predicted by similarity-weighted voting over the top-$k$ neighbors. Second, we train a 2-layer lightweight multilayer perceptron (MLP) on the same frozen embeddings under the identical cross-validation splits. We report binary tension classification and cascade evaluation under video-level 4-fold and stratified 5-fold cross-validation.
\section{Validation}
\label{sec:experiments}

\subsection{Implementation Details}

\paragraph{\textbf{Video Backbone Baselines.}}
We use SlowFast R50 4$\times$16~\cite{feichtenhofer2019slowfast}, TimeSformer~\cite{bertasius2021timesformer}, Video Swin-3D~\cite{liu2022videoswin}, and SurgMotion~\cite{wu2026surgmotion} as frozen video backbone baselines.
SlowFast uses a Kinetics-400~\cite{kay2017kinetics} pretrained checkpoint and extracts features from 32 input frames with a $224 \times 224$ crop.
TimeSformer uses a Kinetics-600~\cite{carreira2018kinetics600} pretrained checkpoint; for evaluation, we uniformly sample 16 frames and resize them to $224 \times 224$.
Video Swin-3D uses the base version and a Kinetics-400~\cite{kay2017kinetics} checkpoint with ImageNet-22K pretraining and processes 32 frames at $224 \times 224$ resolution.
SurgMotion~\cite{wu2026surgmotion} is included as a surgical video foundation model baseline.
All baseline backbones are frozen, and we extract one clip-level embedding from each method for downstream classification.

\vspace{-2mm}
\paragraph{\textbf{\methodname model.}}
The framework is initialized from a few-shot action recognition checkpoint pretrained on only a 10K-video subset of Kinetics-400~\cite{kumar2025trokens,zhu2018compound}, rather than the full Kinetics used by the video backbone baselines (as shown in Table~\ref{tab:model_complexity}).
Each clip is represented by $T=8$ frames at $224 \times 224$ resolution.
We track $M=529$ points per clip using the frozen CoTrackerV3~\cite{karaev2025cotracker3}, initialized on a uniform $23\times23$ grid in the first frame. The resulting trajectories are encoded by three complementary branches: (i) a semantic appearance branch based on DINOv2 ViT-B/14~\cite{oquab2023dinov2}, (ii) an intra-trajectory histogram-of-displacements branch with $K=32$ bins, and (iii) an inter-trajectory motion branch based on all-pair relative positions. Each branch produces $D=768$-dimensional features, which are fused for subsequent spatio-temporal modeling.

\vspace{-2mm}
\paragraph{\textbf{Downstream classifiers.}}
We evaluate frozen clip-level features with two light-weight classifiers:  $k$-NN and a simple MLP (256 units, ReLU, dropout 0.3, AdamW). For $k$-NN, features are z-scored using train-fold statistics, 
$\ell_2$-normalized, and classified by similarity-weighted voting  ($1/(1-sim+\epsilon)$ over cosine neighbors).
We report main results at $k{=}3$.
For the MLP, we apply z-score normalization only, using statistics estimated on the training subset of eac\vspace{-2mm}
\subsection{Evaluation Details}
\label{sec:eval_details}

We evaluate all methods on clip-level frozen features extracted from four video backbones and our trajectory-based framework with \textbf{3{,}593} valid clips (7 videos).

\vspace{-1.5mm}
\paragraph{\textbf{Cross-validation protocols.}}
We use two complementary evaluation protocols.
\textbf{(i) Video-grouped $\rho$-fold validation} evaluates cross-video
generalization by assigning all clips from the same source video
exclusively to either the training or test set within each fold.
We use four-fold ($\rho{=}4$) stratified group cross-validation because the seven source videos exhibit substantially different class distributions. Compared with leave-one-video-out evaluation, grouping the videos into four folds improves the coverage of the tension categories in the test folds while maintaining strict source-video separation.
\textbf{(ii) Stratified clip-level $\rho$-fold validation} combines all
3{,}593 valid clips from the seven videos and performs five-fold
cross-validation ($\rho{=}5$), with stratified splits that approximately
preserve the per-class distribution in each fold.

\vspace{-3mm}
\paragraph{\textbf{Classification tasks.}}
We report results for two task formulations:
\textbf{Binary tension detection} distinguishes between non-tension and tension categories on all 3{,}593 clips.\textbf{Cascade tension recognition} addresses the full 5-way problem (levels~0--4) on all 3{,}593 clips via a two-stage pipeline: Stage~1 performs binary tension detection; Stage~2 performs 4-class level grading, trained only on tension-present training clips.

\vspace{-3mm}
\paragraph{\textbf{Validation Metrics.}}
Given the imbalanced class distribution, we use macro-F1 as the primary evaluation metric, as it averages the F1 scores across different classes and gives each class equal importance. We additionally report accuracy and balanced accuracy for a more comprehensive comparison.

\subsection{Results}

\subsubsection{\textbf{Model complexity and pretraining scale.}} Table~\ref{tab:model_complexity} compares the representation complexity and pretraining scale of \methodname with the video-based baselines. While the video backbones were pretrained by optimizing their full encoders, \methodname uses frozen DINOv2 and CoTrackerV3 models for offline feature and trajectory extraction and trains only an 18M-parameter trajectory encoder. All encoders are subsequently frozen for SurgTension evaluation. After offline feature and trajectory extraction, \methodname uses representations extracted from only $8{\times}1$ frames and requires 35 GFLOPs for trajectory encoding. This is comparable to SlowFast at 36 GFLOPs, while using approximately half the trainable parameters and one quarter of the input frames. Its trajectory encoder is also substantially lighter than the video encoders of Video Swin-3D, TimeSformer, and SurgMotion, and is pretrained on a 10K-video subset of Kinetics-400. Thus, \methodname provides a compact and efficient representation for tissue tension recognition.

\vspace{-5mm}
\subsubsection{{\textbf{Comparison with Video backbone baselines.}}}

Under stratified clip-level validation of binary tension classification in Table~\ref{tab:classifier_comparison}, all methods perform similarly, with macro-F1 scores between 82.2\% and 85.3\%. \methodname achieves 84.7\% with $k$-NN and 85.3\% with the MLP, the latter matching the best result. Trajectory-based features remain competitive under the more challenging video-grouped evaluation. With $k$-NN, \methodname obtains the best accuracy, balanced accuracy, and macro-F1, reaching 82.4\%, 76.4\%, and 75.6\%, respectively. With the MLP, it further achieves the highest balanced accuracy of 76.8\% (tied with Video Swin-3D) and a competitive macro-F1 of 75.8\%. These results suggest that trajectory-based features can support cross-video generalization.

\begin{table}[!t]
\centering
\caption{Comparison of representation-encoder complexity and pretraining scale. Trainable parameters and pretraining refers to the encoders optimized during their representation-learning stages. For \methodname, DINOv2 and CoTrackerV3 remain frozen and performed offline; the reported parameters, GFLOPs, and pretraining scale cover only the trajectory encoder after offline feature and trajectory extraction.}
\label{tab:model_complexity}
\scriptsize
\setlength{\tabcolsep}{2.2pt}
\resizebox{\columnwidth}{!}{%
\begin{tabular}{lccccc}
\toprule
\multirow{2}{*}{Method} &
\multirow{2}{*}{Trainable Params} & 
\multirow{2}{*}{Frames} &
\multirow{2}{*}{Encoder GFLOPs} &
\multicolumn{2}{c}{Pretraining} \\
\cmidrule(lr){5-6}
& & & & Dataset & Scale \\
\midrule

SlowFast~\cite{feichtenhofer2019slowfast} R50 $4{\times}16$
& 34M & $32{\times}1$
& 36 & K400~\cite{kay2017kinetics} & 240K videos \\

Video Swin-3D~\cite{liu2022videoswin}
& 88M & $32{\times}1$
& 282 & IN-21K$\rightarrow$K400~\cite{kay2017kinetics}
& 14M img. + 240K vid. \\

TimeSformer~\cite{bertasius2021timesformer} DivST
& 122M & $16{\times}1$
& 362 & IN-21K$\rightarrow$K600~\cite{carreira2018kinetics600}
& 14M img. + 392K vid. \\

SurgMotion~\cite{wu2026surgmotion} ViT-L
& 304M & $64{\times}1$
& 5,785 & SurgMotion-15M~\cite{wu2026surgmotion}
& 15M images \\

\midrule
\textbf{\methodname}
& 18M & $8{\times}1$
& 35 & K400-10K~\cite{kay2017kinetics,zhu2018compound}
& 10K videos \\

\bottomrule
\end{tabular}}
\raggedright
\scriptsize
\end{table}

\begin{table*}[!t]
\centering
\scriptsize
\caption{Comparison of KNN ($k=3$) and MLP classifiers on frozen features for binary tension classification and end-to-end cascade tension recognition. Stratified evaluation uses clip-level stratified splits, whereas video-grouped evaluation assigns all clips from the same surgical video exclusively to the same fold, ensuring strict source-video separation. All metrics are reported in \%.}
\label{tab:classifier_comparison}

\setlength{\tabcolsep}{3.5pt}
\renewcommand{\arraystretch}{0.95}

\begin{tabular}{lllcccccc}
\toprule
Task & Method & Classifier &
\multicolumn{3}{c}{Stratified} &
\multicolumn{3}{c}{Video-grouped} \\
\cmidrule(lr){4-6}\cmidrule(lr){7-9}
&&
& Acc. & BalAcc. & M-F1
& Acc. & BalAcc. & M-F1 \\
\midrule

\multirow{10}{*}{Binary}
& \multirow{2}{*}{SlowFast~\cite{feichtenhofer2019slowfast}}
& KNN & 87.6 & 80.4 & 82.2 & 81.1 & 72.7 & 73.6 \\
&& MLP & 87.7 & 84.3 & 83.7 & 79.7 & 75.9 & 74.3 \\

& \multirow{2}{*}{TimeSformer~\cite{bertasius2021timesformer}}
& KNN & 89.7 & 83.6 & 85.2 & 81.9 & 72.4 & 73.9 \\
&& MLP & 88.6 & 85.1 & 84.8 & 83.0 & 76.2 & 76.6 \\

& \multirow{2}{*}{Video Swin-3D~\cite{liu2022videoswin}}
& KNN & 89.7 & 83.6 & 85.3 & 81.5 & 73.1 & 74.0 \\
&& MLP & 88.9 & 85.7 & 85.3 & 81.9 & 76.8 & 76.2 \\

& \multirow{2}{*}{SurgMotion~\cite{wu2026surgmotion}}
& KNN & 89.3 & 84.0 & 85.0 & 79.0 & 71.5 & 71.4 \\
&& MLP & 88.8 & 85.4 & 85.1 & 79.4 & 75.9 & 74.1 \\

& \multirow{2}{*}{\textbf{\methodname}}
& KNN & 89.2 & 83.5 & 84.7 & 82.4 & 76.4 & 75.6 \\
&& MLP & 88.9 & 85.2 & 85.3 & 81.2 & 76.8 & 75.8 \\

\midrule

\multirow{10}{*}{Cascade}
& \multirow{2}{*}{SlowFast~\cite{feichtenhofer2019slowfast}}
& KNN & 59.6 & 52.9 & 53.7 & 38.9 & 31.5 & 31.9 \\
&& MLP & 59.6 & 55.7 & 55.4 & 40.8 & 33.8 & 33.8 \\

& \multirow{2}{*}{TimeSformer~\cite{bertasius2021timesformer}}
& KNN & 66.6 & 60.5 & 60.8 & 41.1 & 33.9 & 34.5 \\
&& MLP & 63.8 & 57.6 & 58.4 & 43.4 & 36.0 & 36.2 \\

& \multirow{2}{*}{Video Swin-3D~\cite{liu2022videoswin}}
& KNN & 67.3 & 62.6 & 62.7 & 41.8 & 32.7 & 33.0 \\
&& MLP & 65.5 & 62.8 & 61.2 & 45.6 & 38.2 & 37.8 \\

& \multirow{2}{*}{SurgMotion~\cite{wu2026surgmotion}}
& KNN & 66.5 & 61.5 & 61.9 & 40.8 & 32.2 & 31.9 \\
&& MLP & 65.5 & 60.8 & 60.5 & 43.4 & 34.1 & 33.4 \\

& \multirow{2}{*}{\textbf{\methodname}}
& KNN & 64.6 & 59.3 & 59.4 & 40.0 & 31.5 & 31.7 \\
&& MLP & 63.4 & 59.0 & 58.9 & 42.2 & 35.6 & 35.8 \\

\bottomrule
\end{tabular}
\vspace{-4mm}
\end{table*}

Table~\ref{tab:classifier_comparison} also reports end-to-end two-stage cascade results, where errors in tension detection can propagate to tension-level grading, making the overall recognition task more challenging. With $k$-NN, \methodname achieves macro-F1 scores of 59.4\% and 31.7\% under stratified and video-grouped validation, respectively, outperforming SlowFast under the stratified protocol and remaining comparable to several video backbones under video-grouped evaluation. With the MLP, it obtains 58.9\% and 35.8\% macro-F1, outperforming SlowFast under both protocols and approaching TimeSformer under video-grouped evaluation, with only a 0.4-point gap. Although Video Swin-3D achieves the strongest overall cascade performance, \methodname remains competitive across classifiers and evaluation protocols.

\begin{figure}[t]
    \centering
    \includegraphics[width=0.99\linewidth]{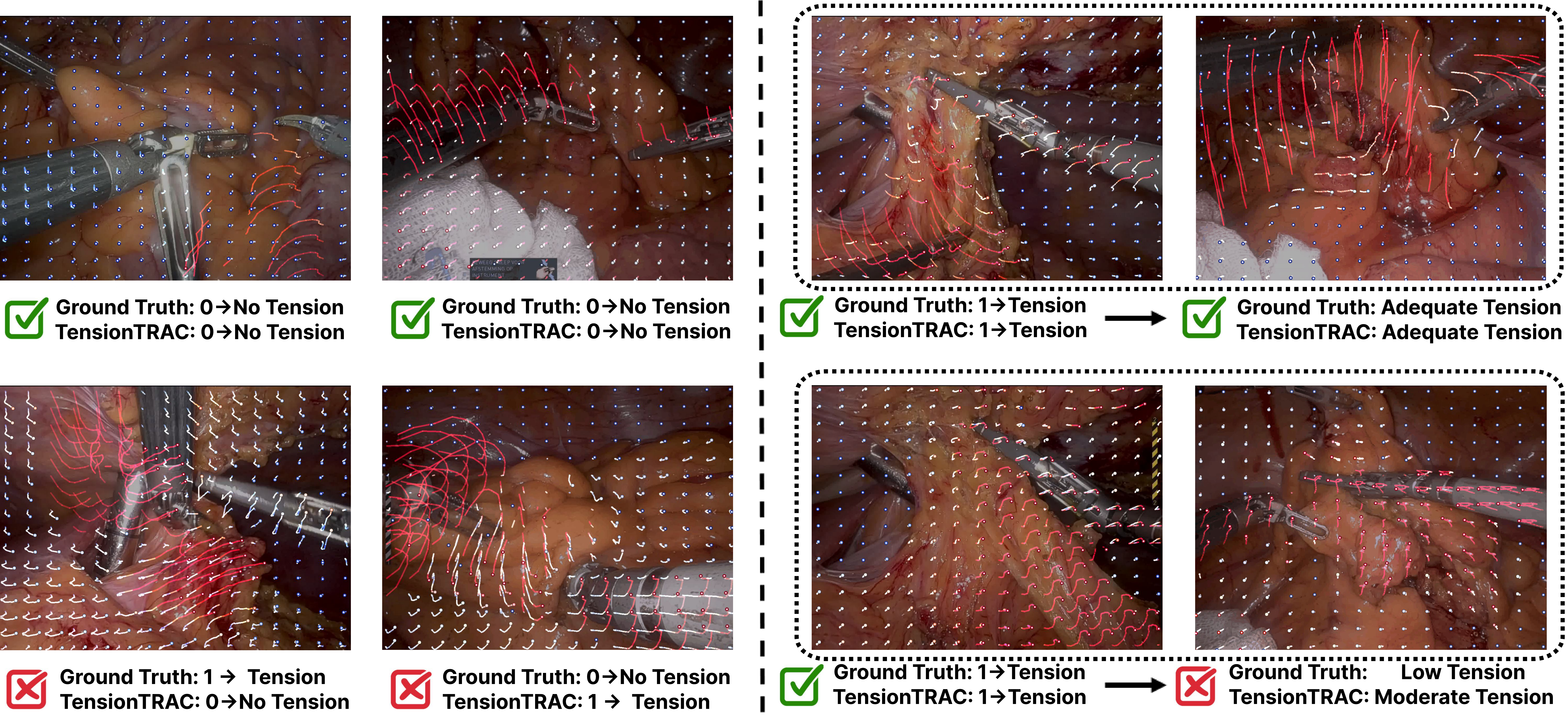}
    \caption{Qualitative results of \methodname for tissue tension recognition. The examples present predictions for binary tension classification (left) and cascade two-stage evaluation (right). Sparse trajectories visualize local tissue and instrument motion in tension-present clips, while failure cases arise from camera, instrument, or other non-tension-related motion, as well as confusion between neighboring tension levels.}
    \label{fig:qualitative_results}
    \vspace{-3mm}
\end{figure}

\vspace{-2mm}
\paragraph{\textbf{Qualitative Results.}}
Figure~\ref{fig:qualitative_results} shows predictions of \methodname with trajectory visualizations. Correctly classified tension clips generally exhibit structured local tissue deformation, whereas non-tension clips mainly contain limited or tension-unrelated motion. Failure cases reflect several challenges. Occlusion and visual disturbances such as smoke or fluid can reduce the visibility and reliability of trajectories extracted by the frozen point tracker. Visibility masking and visibility-weighted aggregation reduce the influence of invalid or low-visibility trajectory positions, although severe visual degradation may still affect the representation. In contrast, camera motion, instrument motion, and unrelated tissue deformation can produce valid but tension-irrelevant trajectories resembling traction-induced motion. False negatives may also occur when tension-induced deformation is subtle or highly localized. The cascade results further reveal confusion between adjacent tension levels, such as Low being misclassified as Moderate. These observations highlight the challenges of robust trajectory extraction and separating tension-induced deformation from other motion.

\section{Conclusion}
\label{sec:conclusion}

We introduce tissue tension recognition as a new surgical video understanding task and present \dataname, a clinically annotated and expert-reviewed dataset for tension recognition. We further propose TensionTRAC, a lightweight trajectory-based framework that captures local tissue motion and deformation by jointly modeling intra-trajectory dynamics, inter-trajectory interactions, and point-aligned semantic features. Experiments on binary tension classification and end-to-end five-class cascade recognition demonstrate that TensionTRAC achieves competitive performance while using fewer input frames, parameters, and pretraining data than several strong video backbones. The results support sparse trajectory modeling as an efficient representation for tissue tension analysis, while also highlighting the remaining challenges of fine-grained tension grading and generalization across surgical videos. We hope that SurgTension and TensionTRAC will facilitate future research on clinically meaningful surgical video understanding and objective surgical performance assessment.

\clearpage
\section*{Acknowledgements}
This work was supported by the Twente University Radboudumc Opportunities (TURBO) programme through the Surgical VideoAI: Automation of Video-Based Surgical Skill Assessment in Rectal Cancer project. The TURBO programme is part of HealthTech Nexus, the strategic partnership between the University of Twente and Radboudumc. Computational resources were provided by the University of Twente High-Performance Computing infrastructure and the University of Groningen Hábrók computing cluster.


\bibliographystyle{splncs04}
\bibliography{references}

\end{document}